\pdfoutput=1
\documentclass[10pt,twocolumn,letterpaper]{article}

\usepackage{cvpr}              

\usepackage{times}
\usepackage{epsfig}
\usepackage{graphicx}
\usepackage{amsmath}
\usepackage{amssymb}
\usepackage{booktabs} 
\usepackage{multirow}
\usepackage{makecell}
\usepackage{amsfonts}
\usepackage{nicefrac}
\usepackage{color}
\usepackage{float}
\usepackage{caption}
\usepackage{lipsum}
\usepackage[table]{xcolor}

\definecolor{cvprblue}{rgb}{0.21,0.49,0.74}
\usepackage[pagebackref,breaklinks,colorlinks,citecolor=cvprblue]{hyperref}

\title{PG-NeuS: Robust and Efficient Point Guidance for Multi-View Neural Surface Reconstruction}

\author{Chen Zhang\textsuperscript{\rm 1} \and Wanjuan Su\textsuperscript{\rm 1} \and Qingshan Xu\textsuperscript{\rm 2} \and Wenbing Tao\textsuperscript{\rm 1,}\thanks{Corresponding author.} \\
\vspace{-1cm}
\and
\textsuperscript{1}Huazhong University of Science and Technology \\
\textsuperscript{2}Nanyang Technological University\\
{\tt\small \textsuperscript{1}\{zhangchen\_, suwanjuan, wenbingtao\}@hust.edu.cn}
{\tt\small \textsuperscript{2}\{qingshan.xu\}@ntu.edu.sg}
}
\begin{document}
\maketitle
\begin{abstract}
Recently, learning multi-view neural surface reconstruction with the supervision of point clouds or depth maps has been a promising way. However, due to the underutilization of prior information, current methods still struggle with the challenges of limited accuracy and excessive time complexity. In addition, prior data perturbation is also an important but rarely considered issue. To address these challenges, we propose a novel point-guided method named PG-NeuS, which achieves accurate and efficient reconstruction while robustly coping with point noise. Specifically, aleatoric uncertainty of the point cloud is modeled to capture the distribution of noise, leading to noise robustness. Furthermore, a Neural Projection module connecting points and images is proposed to add geometric constraints to implicit surface, achieving precise point guidance. To better compensate for geometric bias between volume rendering and point modeling, high-fidelity points are filtered into a Bias Network to further improve details representation. Benefiting from the effective point guidance, even with a lightweight network, the proposed PG-NeuS achieves fast convergence with an impressive 11x speedup compared to NeuS. Extensive experiments show that our method yields high-quality surfaces with high efficiency, especially for fine-grained details and smooth regions, outperforming the state-of-the-art methods. Moreover, it exhibits strong robustness to noisy data and sparse data.
\end{abstract}    
\section{Introduction}
Surface reconstruction from multiple calibrated views is a long-standing problem in computer vision and graphics. Recently, neural surface reconstruction with volume rendering \cite{wang2021neus, yariv2021volume, darmon2022improving, wang2022hf, cai2023neuda} is highly valued for its potential to improve and simplify multi-view 3D reconstruction. These methods introduce the Signed Distance Function (SDF) to express opacity in the radiance field, enabling indirect SDF optimization during rendering. Impressively, photo-realistic color appearance and more complete geometry are modeled compared to the traditional framework. However, due to the lack of explicit geometric constraints, geometric bias in volume rendering \cite{fu2022geo, zhang2023towards} and excessive optimization time become its fly in the ointment.  

\begin{figure}[t]
	\centering
	\includegraphics[width=\linewidth]{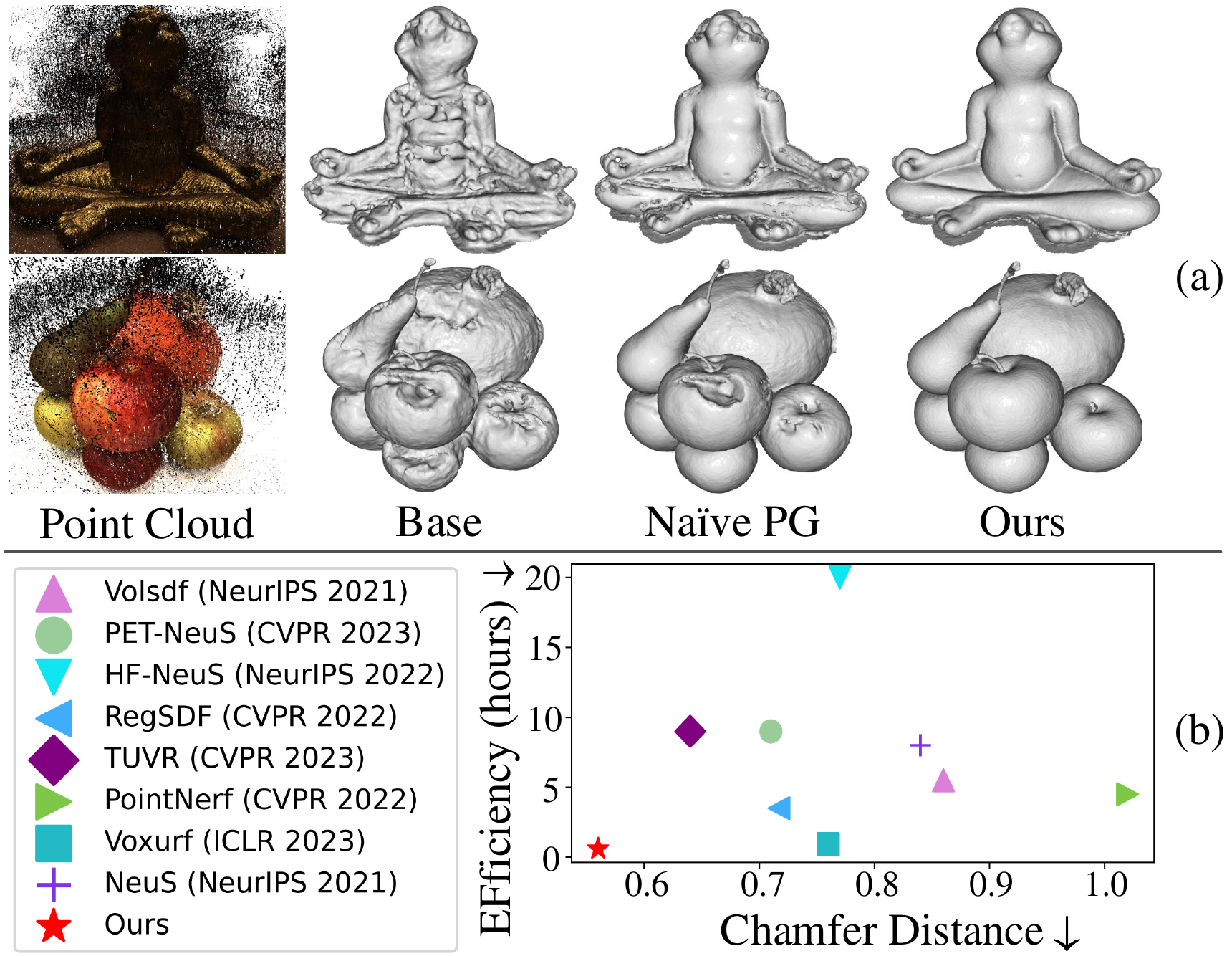}
	\vspace{-0.6cm}
	\caption{(a) Reconstruction results of our method with high-intensity noisy data. "Base" is the baseline for our method without point cloud inputs. "Naive PG" is a naive point guidance mechanism as explained in our Preliminaries Section. (b) Comparison with SOTA methods on accuracy and efficiency.}
	\label{fig0}
	\vspace{-0.4cm}
\end{figure} 

Some methods introduce additional geometric priors to explicitly guide the implicit surface, aiming to improve the reconstruction accuracy. For example, MonoSDF \cite{yu2022monosdf} and MVSDF \cite{zhang2021learning} leverage depth maps to guide the learning of SDF network. Recognizing the redundant geometric information in depth maps, some approaches, such as RegSDF \cite{zhang2022critical} and TUVR \cite{zhang2023towards}, opt for point clouds as a more reliable and concise alternative for supervisory signals, leading to an improvement in the geometric quality of reconstructions. However, these methods still face several key challenges in achieving high-quality point guidance.

1) \textbf{Noise distribution across different regions are difficult to estimate and eliminate.} Point noise stands out as a crucial factor influencing the quality of point guidance. Its characteristic of varying with the regions makes it difficult to handle well. RegSDF \cite{zhang2022critical} introduces a uniform smoothing term for all points to mitigate the impact of noise. However, such a manner neglects the characteristic of noise and also weakens the strong support offered by precise points.

2) \textbf{Distinguishing high-fidelity data is challenging, resulting in underutilization of high-fidelity information.} Utilizing high-fidelity data for precise reconstruction is expected and necessary for point guidance. TUVR \cite{zhang2023towards} takes confidence values output from the MVS network as an indicator of point reliability, assigning higher weights to points with greater confidence. However, its reconstruction accuracy is still limited due to the unreliability of confidence maps as adjuncts to MVS networks. Additionally, this manner is unable to handle data captured by depth cameras. To achieve high-quality point guidance, point reliability needs to be assessed during the geometry optimization process.   

3) \textbf{Accuracy and efficiency are hard to balance.} Despite using priors as guidance, these methods still need several hours or even longer for optimization, thirsting for an efficient reconstruction solution. In fact, the trade-off between efficiency and accuracy is a common challenge, as prioritizing one often comes at the expense of the other. Some methods use complex networks in an attempt to achieve a balance, but with limited effectiveness. In contrast, our method demonstrates that integrating a robust point guidance mechanism allows for achieving superior accuracy and efficiency, even using lightweight networks.

To address the above challenges, we present PG-NeuS, a universal framework with robust point guidance mechanisms to enhance and accelerate neural surface reconstruction. Different from previous point-guided methods, we model the SDF value of each point as a Gaussian distribution. Aleatoric uncertainty estimation is introduced into point modeling to capture the distribution of inherent noise and estimate point reliability. Furthermore, we propose a Neural Projection module to achieve accurate point guidance. In this process, the point data undergoes two projections: first in space onto the implicit surface and then onto the image. Photometric consistency constraint is imposed on the projected positions, serving as the regularization of SDF. Finally, a filter operation is performed utilizing the modeled uncertainty, and high-fidelity points are fed into a Bias Network to compensate for the geometric bias between volume rendering and point modeling. We conduct experiments on the DTU \cite{jensen2014large} and BlendedMVS \cite{yao2020blendedmvs} datasets for evaluations. The results demonstrate the robustness of our method across some challenging tasks, particularly with noisy point data (see Figure \ref{fig0}.a) and sparse point data. Benefiting from the outstanding point guidance, superior reconstruction accuracy and efficient convergence are achieved with lightweight networks, enabling an impressive 33.3\% improvement in accuracy and an 11x speedup compared to NeuS \cite{wang2021neus} without masks (see Figure \ref{fig0}.b). In summary, our contributions are as follows:
\begin{itemize}
	\item By employing uncertainty modeling to capture inherent noise and a Neural Projection module to regularize the SDF, accurate reconstruction and strong noise robustness are achieved.
	\item With the estimated point reliability, the Bias Network compensates for the geometric bias between volume rendering and point modeling, leading to more precise details representation.
	\item Benefiting from effective point guidance, lightweight networks are allowed to achieve efficient convergence while maintaining high accuracy. 
\end{itemize}

\section{Related Work}
\noindent\textbf{Traditional multi-view reconstruction.} Traditional multi-view reconstruction typically involves two steps: fusing depth maps obtained from MVS \cite{zhang2023geomvsnet, su2023efficient, ding2022transmvsnet} or depth cameras into a point cloud, followed by employing point cloud surface reconstruction algorithms \cite{sitzmann2020implicit, ma2021neural} to generate a mesh. Classical screened Poisson Surface Reconstruction (sPSR) \cite{kazhdan2013screened} is a widely used technique to generate watertight surfaces from point clouds. Although the traditional pipeline has made successive breakthroughs in reconstruction quality due to accumulated technical progress, there are still challenges of sensitivity to point noise and density.

\begin{figure*}[t]
	\centering
	\includegraphics[width=\textwidth]{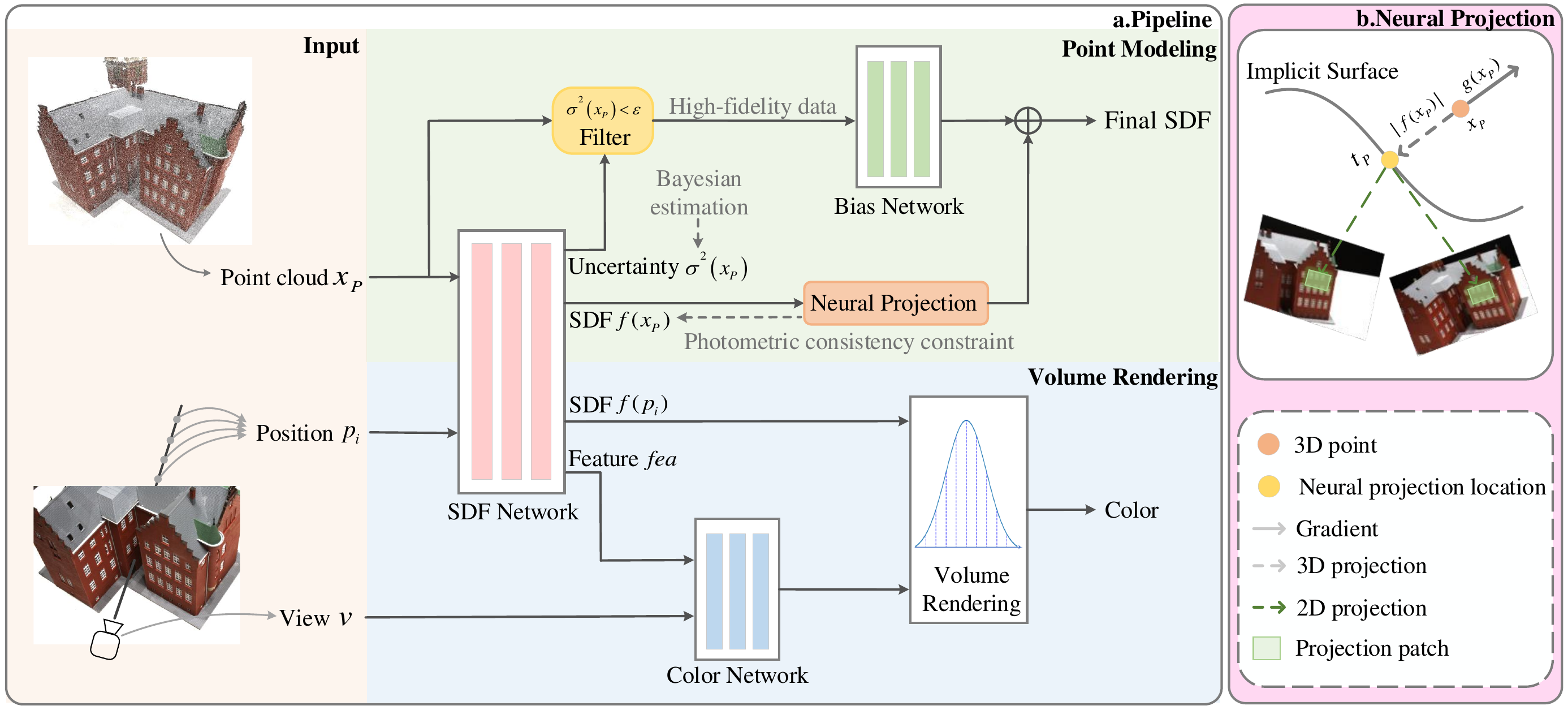}
	\vspace{-0.6cm}
	\caption{(a) The pipeline of Point-NeuS. Our network integrates point modeling with volume rendering through the SDF network, enhancing scene representation. NeuS serves as the base framework for the volume rendering part, while the point modeling part focuses on improving the quality of surface guidance. To this end, uncertainty estimation and a Neural Projection module are designed to attenuate the effect of noise, allowing accurate reconstruction. Additionally, the Bias network further learns to compensate for the geometric bias between volume rendering and point modeling for high-fidelity data. (b) The mechanism of Neural Projection module.}
	\label{fig1}
	\vspace{-0.2cm}
\end{figure*}

~\\
\noindent\textbf{Multi-view neural surface reconstruction.}
Recently, with the development of differentiable rendering \cite{chen2019learning, aliev2020neural, bangaru2022differentiable, tancik2022block, yang2021learning}, neural surface reconstruction \cite{yariv2020multiview, yang2022neumesh,qiu2023looking,azinovic2022neural,zhu2022nice,xiang2021mobile3dscanner} is valued for its single-stage optimization and strong representation potential. Nerf \cite{mildenhall2021nerf} as a pioneer triggers plenty of works \cite{martin2021nerf, barron2021mip, pan2022activenerf, chen2021mvsnerf, shen2022conditional} to represent scenes within the volume rendering framework. To improve geometric quality, VolSDF \cite{yariv2021volume} and NeuS \cite{wang2021neus} design new weight functions involving SDF for color accumulation in the rendering. Compared to traditional pipelines, these methods not only render photo-realistic appearances but also reconstruct more complete surfaces, especially for non-Lambertian surfaces. However, geometric bias is still an urgent problem due to discrete sampling and other factors \cite{fu2022geo, zhang2023towards}. Additionally, lengthy optimization also hinders their broader application. Subsequent methods \cite{wang2022hf, wang2023pet, sun2022direct} try to solve these problems, but still struggle to strike an accuracy-efficiency balance. Impressively, some methods use voxel-based representation to enhance training efficiency, such as Voxurf \cite{wu2022voxurf} and Vox-Surf \cite{li2022vox}, making significant strides in achieving this equilibrium.

\noindent\textbf{Geometry supervision for multi-view neural reconstruction.} To achieve high-quality reconstruction, some methods \cite{jiang2023depth, zhu2023vdn} introduce additional geometric information to explicitly guide the geometry modeling. MonoSDF \cite{yu2022monosdf} employs monocular normal and depth cues to supervise the SDF network, while MVSDF \cite{zhang2021learning} works with depth information from an MVS network. Related to our method, RegSDF \cite{zhang2022critical} and TUVR \cite{zhang2023towards} use point clouds as supervisory signals, enhancing scene representation while accounting for point noise. RegSDF smooths all data by constraining the second derivative, which overlooks noise variation across regions. On the other hand, TUVR simply relies on confidence value output by the MVS network to avoid this problem. In contrast, our method introduces uncertainty to capture the distribution of noise, allowing surface optimization and point reliability assessment to be mutually reinforcing. Although the additional geometry priors accelerate the convergence of these methods, they still require several hours to optimize a scene. Thus, an efficient implementation is still desperately needed. Additionally, point clouds are also used by some methods to render high-quality images, such as Point-NeRF \cite{xu2022point}, NPBG \cite{aliev2020neural} and SNP \cite{zuo2022view}, but they do not focus on implicit surface quality.

\section{Preliminaries}
\label{sec:Preliminaries}
\noindent\textbf{Reconstruction from volume rendering.} NeuS \cite{wang2021neus} is a very popular method for neural implicit surface reconstruction. It represents the surface as the zero-level set of SDF $\mathcal{S}=\{x\in {{\mathbb{R}}^{3}}|f\left( x \right)=0\}$. In rendering, a ray emitting from the camera center $o$ along the direction $d$ that passes through a pixel can be expressed as $\{p(t)=o+td|t>0\}$. The rendered color $\hat{C}$ of this pixel is accumulated along the ray with $N$ discrete sampled points.
\begin{equation}
	\label{eqn1}
	\hat{C}=\sum\limits_{i=1}^{N}{{{T}_{i}}{{\alpha }_{i}}{{c}_{i}}}, \quad {{T}_{i}}=\prod\limits_{j=1}^{i-1}{(1-{{\alpha }_{j}})}
\end{equation}
where ${\alpha }_{i}$ is the discrete opacity, and ${T}_{i}$ is the accumulated transmittance. To reduce the estimation bias of SDF, NeuS defines the opacity as follows:
\begin{equation}
	\label{eqn2}
	{{\alpha }_{i}}=\max (\frac{{{\Phi }_{s}}(f(p({{t}_{i}})))-{{\Phi }_{s}}(f(p({{t}_{i+1}})))}{{{\Phi }_{s}}(f(p({{t}_{i}})))},0)
\end{equation}
Here, ${{\Phi }_{s}}(x)={{(1+{{e}^{-sx}})}^{-1}}$ is the Sigmoid function, where the $s$ value is a trainable parameter.

~\\
\noindent\textbf{Naive point guidance.} Given a point cloud $\{P|{{x}_{P}}\in P\}$, a straightforward guidance for neural surface reconstruction is to assume that all points are distributed over the implicit surface, such that the SDF network can be effectively supervised by $\{f({{x}_{P}})=0\}$. However, in reality, the point cloud distribution varies around the implicit surface, leading to varying degrees of noise across different regions. The key to achieving high-quality reconstruction lies in effectively removing noise and leveraging high-fidelity points to enhance detail expression.

\section{Method}
Given a set of RGB images with known camera parameters, our aim is to accurately reconstruct geometry within the volume rendering framework. To counteract the inherent geometric bias in the rendering, a point cloud is additionally introduced as the guidance for the implicit surface to achieve high-quality and efficient reconstruction. The pipeline of our method is illustrated in Figure \ref{fig1}.a.

\subsection{Point Cloud Generation}
Point clouds can be obtained by fusing depth maps obtained from MVS or depth cameras. For simplicity, we use TransMVSNet \cite{ding2022transmvsnet} with robust generalization ability to generate a depth map for each view. Proper geometric consistency filtering is performed in the fusion process to produce relatively accurate and clean point clouds.

\subsection{Uncertainty Estimation}
In Bayesian deep learning \cite{kendall2017uncertainties, lakshminarayanan2017simple}, the inherent noise of the observed data is modeled as aleatoric uncertainty. Inspired by this, we learn the uncertainty for the input point cloud to capture the distribution of noise. For each point ${x}_{P}$ in point set $P$, its SDF value is modeled as a Gaussian distribution rather than a single value, where the predicted variance can reflect the aleatoric uncertainty of ${x}_{P}$. Considering that the noise varies across regions, each point is allowed to learn a different variance. As a result, the model provides a large variance for points with high noise, rather than collapsing to a trivial solution. The learning of heteroscedastic aleatoric uncertainty can well help our method capture complex noise for better geometry optimization. 

Specifically, the SDF value of a point ${x}_{P}$ in $P$ is output as a Gaussian distribution parameterized by mean $f\left( {{x}_{P}} \right)$ and variance ${{\widetilde{\sigma }}^{2}}\left( {{x}_{P}} \right)$. We take the output of the SDF network as the mean and add an additional header to model the variance.
\begin{equation}
	\label{eqn3}
	f\left( {{x}_{P}} \right),{{\sigma }^{2}}\left( {{x}_{P}} \right), fea=ML{{P}_{sdf}}({{x}_{P}})
\end{equation}
$fea$ is the feature vector fed into the color network, but we do not use it for point modeling. Softplus function is further adopted to ${\sigma }^{2}$ to produce a valid variance value. Then a minimum variance constraint $\sigma _{0}^{2}$ is added for all points.
\begin{equation}
	\label{eqn4}
	\tilde{\sigma}^2\left(x_P\right)=\sigma_0^2+\log \left(1+\exp \left(\sigma^2\left(x_P\right)\right)\right)
\end{equation}
To optimize the model, we assume that the SDF values of all points fluctuate around the 0 value. This assumption is reasonable since the ideal point cloud is distributed over a 0-value surface. Therefore, following the theoretical guidance of Bayesian deep learning \cite{kendall2017uncertainties}, the negative log-likelihood of a point set is optimized in the regression as follows:
\begin{equation}
	\label{eqn5}
	\begin{split}
		&\qquad\quad {{\mathcal{L}}_{usdf}} =-\frac{1}{{{D}_{P}}}\sum\limits_{{{x}_{P}}\in P}{\log {{p}_{\theta }}({f}_{\varepsilon}({{x}_{P}}))} \\ 
		& =\frac{1}{{{D}_{P}}}\sum\limits_{{{x}_{P}}\in P}{\frac{\left\| {{f}_{\varepsilon}}({{x}_{P}})-f\left( {{x}_{P}} \right) \right\|_{2}^{2}}{2{{\widetilde{\sigma }}^{2}}\left( {{x}_{P}} \right)}}+\frac{\log {{\widetilde{\sigma }}^{2}}\left( {{x}_{P}} \right)}{2}
	\end{split}
\end{equation}
where ${D}_{P}$ represents the number of points in the set $P$, and ${f}_{\varepsilon}({{x}_{P}})$ denotes the ideal SDF value for ${x}_{P}$,  which is set to 0. In the first term in the loss, large value of ${{\widetilde{\sigma }}^{2}}\left( {{x}_{P}} \right)$ reduces the importance assigned to a point with high noise, thereby mitigating adverse effects. The second term is a regularization term to avoid predicting infinite variance for all points. 

\subsection{Bias Network}
To achieve superior point guidance, it is not only essential to eliminate noise-related effects, but also to utilize high-fidelity data for more precise reconstruction. For each point ${x}_{P}$ with a predicted variance ${{\widetilde{\sigma }}^{2}}\left( {{x}_{P}} \right)$, a larger ${{\widetilde{\sigma }}^{2}}\left( {{x}_{P}} \right)$ represents a higher uncertainty and also a lower reliability for ${x}_{P}$. To focus on learning precise geometric representations with reliable points, we perform a filtering operation to extract the point set ${P}_{s}$ with a variance below the threshold $\varepsilon$, i.e., $\{{{P}_{s}}|{{x}_{P}}\in {{P}_{s}},{{\widetilde{\sigma }}^{2}}\left( {{x}_{P}} \right)<\varepsilon \}$. ${{P}_{s}}$ is further fed to a Bias network to capture the geometric bias between volume rendering and point modeling. This process aims to filter noisy data and enhance the representation of details. With the SDF network in volume rendering taken as the base SDF network, the final SDF value of ${x}_{P}$ is the sum of the bias SDF value ${{f}_{b}}\left( {{x}_{P}} \right)$ and the base SDF value $f({{x}_{P}})$. The whole process is formulated as follows:
\begin{equation}
	\label{eqn6}
	{f}_{b}\left( {{x}_{P}} \right)=ML{{P}_{bias}}({{x}_{P}})
\end{equation}
\begin{equation}
	\label{eqn7}
	{{f}_{final}}\left( {{x}_{P}} \right)=f({{x}_{P}})+{{f}_{b}}\left( {{x}_{P}} \right)
\end{equation}
Furthermore, we design a SDF loss to optimize the model.
\begin{equation}
	\label{eqn8}
	{{\mathcal{L}}_{bias}}=\frac{1}{{{D}_{Ps}}}\sum\limits_{{{x}_{P}}\in {{P}_{s}}}{|{{f}_{\varepsilon }}({{x}_{P}})-{{f}_{final}}({{x}_{P}})|}
\end{equation}
where ${{D}_{Ps}}$ is the number of points in ${P}_{s}$, and ${{f}_{\varepsilon }}({{x}_{P}})$ is the ideal SDF value of ${x}_{P}$, which is set to 0. With the loss, the network learns a more accurate representation of the regions where high-fidelity points are located. 

It is worth noting that this Bias network specifically designed for high-fidelity points works independently of the volume rendering, making our method fundamentally different from another method HF-NeuS \cite{wang2022hf} that is prone to confusion. The latter relies on volume rendering to learn high-frequency information, involving the gradient of SDF field. For more details, please refer to the supplementary.

\subsection{Neural Projection}
In order to enhance the prediction accuracy of SDF values for points while further enforcing geometric constraints on noisy data, a Neural Projection module is designed to regularize the SDF field using photometric consistency. As shown in Figure \ref{fig1}.b, for a point ${x}_{P}$, we use the predicted SDF value $f({{x}_{P}})$ and the gradient $g({{x}_{P}})$ to project it to the nearest neighbor ${t}_{P}$ on the implicit surface $\mathcal{S}$. The gradient $g({{x}_{P}})$ represents the direction vector with the fastest increasing SDF value at ${x}_{P}$. Therefore, ${t}_{P}$ can be obtained by moving ${x}_{P}$ along or against the direction of $g({{x}_{P}})$ by a distance of $|f({{x}_{P}})|$.
\begin{equation}
	\label{eqn9}
	{{t}_{P}}={{x}_{P}}-f\left( {{x}_{P}} \right)\times g({{x}_{P}})/{{\left\| g({{x}_{P}}) \right\|}_{2}}
\end{equation}
$g({{x}_{P}})$ can be also denoted as $\nabla f\left( {{x}_{P}} \right)$, whose direction is always toward the outside of surface $\mathcal{S}$.

In fact, the accuracy of learned projection distance is consistent with the accuracy of constructed implicit surface. To ensure that ${t}_{P}$ is distributed on the real surface, a geometric constraint is necessary. For a small area $s$ on the surface where ${t}_{P}$ is located, its projection patches in different views are supposed to be geometry-consistent. Therefore, we use the photometric consistency constraint from MVS to supervise the position of ${t}_{P}$. The normalization cross-correlation scores of patches between the reference image ${I}_{r}$ and the source image ${I}_{s}$ are calculated according to Equation \ref{eqn10}. The image corresponding to the depth map that generates ${x}_{P}$ is taken as the reference image and the rest as source images. 
\begin{equation}
	\label{eqn10}
	Score({{I}_{r}}(s),{{I}_{s}}(s))=\frac{Cov({{I}_{r}}(s),{{I}_{s}}(s))}{\sqrt{Var({{I}_{r}}(s))Var({{I}_{s}}(s))}}
\end{equation}
where $Cov$ is covariance and $Var$ is variance. ${{I}_{r}}(s)$ and ${{I}_{s}}(s)$ are the patches of $s$ projecting to ${I}_{r}$ and ${I}_{s}$, respectively. The photometric consistency loss ${\mathcal{L}}_{pc}$ is calculated as follows.
\begin{equation}
	\label{eqn11}
	{{\mathcal{L}}_{pc}}=\frac{1}{{m}{{D}_{P}}}\sum\limits_{{{x}_{P}}\in P}{\sum\limits_{i=1}^{m}{(1-Score({{I}_{r}}(s),{{I}_{si}}(s)))}}
\end{equation}
where ${D}_{P}$ is the number of points in the set $P$. $m$ is the number of source images achieving the best scores, which is set to 4. ${{\mathcal{L}}_{pc}}$ in our Neural Projection module imposes a strong geometric constraint on the projection position ${t}_{P}$, indirectly ensuring the accuracy of the predicted SDF value at ${x}_{P}$. This point-to-image constraint is beneficial for learning better SDF representations from noisy data. Therefore, we apply it to the signals output from our base SDF network for enhancing the resistance to noise. Since this process requires derivatives of the networks, this constraint is not imposed on the final SDF for efficiency and avoiding affecting the Bias network.

\begin{table*}[h]
	\centering
	\resizebox{\textwidth}{!}{
		\begin{tabular}{cccccccccccccccc|c|c}
			\toprule[1.2pt]
			Scan ID & 24 & 37 & 40 & 55 & 63 & 65 & 69 & 83 & 97 & 105 & 106 & 110 & 114 & 118 & 122 & Mean & Time \\
			\midrule
			COLMAP\cite{schonberger2016pixelwise} & 0.81 & 2.05 & 0.73 & 1.22 & 1.79 & 1.58 & 1.02 & 3.05 & 1.40 & 2.05 & 1.00 & 1.32 & 0.49 & 0.78 & 1.17 & 1.36 & -\\
			${\text{sPSR}}_{0}$\cite{kazhdan2013screened} & 0.49 & 1.27 & 0.63 & 0.57 & 0.88 & 0.69 & 0.53 & 1.37 & 0.93 & 0.76 & 0.60 & 0.86 & 0.30 & 0.51 & 0.70 & 0.74 & -\\
			${\text{sPSR}}_{7}$\cite{kazhdan2013screened} & \cellcolor{orange!30} 0.47 & 1.10 & 0.44 & \cellcolor{orange!30} 0.34 & 0.81 & 0.69 & 0.53 & 1.21 & 0.91 & 0.68 & 0.51 & 0.70 & 0.30 & \cellcolor{orange!30} 0.40 & 0.48 & \cellcolor{orange!30} 0.64 & -\\
			SIREN\cite{sitzmann2020implicit} & 0.56 & 1.05 & 0.46 & 0.44 & 1.15 & 0.88 & 1.56 & 1.95 & 1.87 & 0.78 & 1.22 & 1.16 & 1.78 & 1.50 & 0.53 & 1.13 & -\\
			\midrule
			\midrule
			VolSDF\cite{yariv2021volume} & 1.14 & 1.26 & 0.81 & 0.49 & 1.25 & 0.70 & 0.72 & 1.29 & 1.18 & 0.70 & 0.66 & 1.08 & 0.42 & 0.61 & 0.55 & 0.86 & 5.5h \\
			NeuS\cite{wang2021neus} & 1.00 & 1.37 & 0.93 & 0.43 & 1.10 & 0.65 & 0.57 & 1.48 & 1.09 & 0.83 & 0.52 & 1.20 & 0.35 & 0.49 & 0.54 & 0.84 & 8h\\
			DVGO\cite{sun2022direct} & 1.83 & 1.74 & 1.70 & 1.53 & 1.91 & 1.91 & 1.77 & 2.60 & 2.08 & 1.79 & 1.76 & 2.12 & 1.60 & 1.80 & 1.58 & 1.85 & \cellcolor{red!30} 15min \\
			HF-NeuS\cite{wang2022hf} & 0.76 & 1.32 & 0.70 & 0.39 & 1.06 & 0.63 & 0.63 & 1.15 & 1.12 & 0.80 & 0.52 & 1.22 & 0.33 & 0.49 & 0.50 & 0.77 & 20h \\
			NeuralWarp\cite{darmon2022improving} & 0.49 & \cellcolor{red!30} 0.71 & \cellcolor{red!30} 0.38 & 0.38 & \cellcolor{orange!30} 0.79 & 0.81 & 0.82 & 1.20 & 1.06 & 0.68 & 0.66 & 0.74 & 0.41 & 0.63 & 0.51 & 0.68 & 5h\\
			PET-NeuS\cite{wang2023pet} & 0.56 & 0.75 & 0.68 & 0.36 & 0.87 & 0.76 & 0.69 & 1.33 & 1.08 & \cellcolor{red!30} 0.66 & 0.51 & 1.04 & 0.34 & 0.51 & 0.48 & 0.71 & 9h \\
			NeuDA\cite{cai2023neuda} & \cellcolor{orange!30} 0.47 & \cellcolor{red!30} 0.71 & 0.42 & 0.36 & 0.88 & \cellcolor{orange!30} 0.56 & 0.56 & 1.43 & 1.04 & 0.81 & 0.51 & 0.78 & 0.32 & 0.41 & 0.45 & 0.65 & - \\
			Voxurf\cite{wu2022voxurf} & 0.72 & 0.75 & 0.47 & 0.39 & 1.47 & 0.76 & 0.81 & 1.02 & 1.04 & 0.92 & 0.52 & 1.13 & 0.40 & 0.53 & 0.53 & 0.76 & 55min\\
			\midrule
			MVSDF\cite{zhang2021learning} & 0.83 & 1.76 & 0.88 & 0.44 & 1.11 & 0.90 & 0.75 & 1.26 & 1.02 & 1.35 & 0.87 & 0.84 & 0.34 & 0.47 & 0.46 & 0.88 & 2.7h\\
			MonoSDF\cite{yu2022monosdf} & 0.66 & 0.88 & 0.43 & 0.40 & 0.87 & 0.78 & 0.81 & 1.23 & 1.18 & \cellcolor{red!30} 0.66 & 0.66 & 0.96 & 0.41 & 0.57 & 0.51 & 0.73 & 5h \\
			Point-NeRF\cite{xu2022point} & 0.87 & 2.06 & 1.20 & 1.01 & 1.01 & 1.39 & 0.80 & \cellcolor{orange!30} 1.04 & 0.92 & 0.74 & 0.97 & 0.76 & 0.56 & 0.90 & 1.05 & 1.02 & 4.5h \\
			RegSDF\cite{zhang2022critical} & 0.597 & 1.410 & 0.637 & 0.428 & 1.342 & 0.623 & 0.599 & \cellcolor{red!30} 0.895 & 0.919 & 1.020 & 0.600 & \cellcolor{red!30} 0.594 & \cellcolor{red!30} 0.297 & 0.406 & \cellcolor{orange!30} 0.389 & 0.717 & 3.5h\\
			TUVR\cite{zhang2023towards} & 0.56 & 0.92 & \cellcolor{orange!30} 0.39 & 0.39 & 0.85 & 0.58 & \cellcolor{orange!30} 0.51 & 1.20 & \cellcolor{orange!30} 0.90 & 0.78 & \cellcolor{orange!30} 0.42 & 0.84 & 0.32 & 0.43 & 0.44 & \cellcolor{orange!30} 0.64 & 9h \\
			Ours & \cellcolor{red!30} 0.338 & 0.822 & 0.403 & \cellcolor{red!30} 0.331 & \cellcolor{red!30} 0.754 & \cellcolor{red!30} 0.541 & \cellcolor{red!30} 0.493 & 1.128 & \cellcolor{red!30} 0.855 & 0.713 & \cellcolor{red!30}  0.367 & \cellcolor{orange!30} 0.606 & \cellcolor{orange!30} 0.302 & \cellcolor{red!30} 0.371 & \cellcolor{red!30} 0.378 & \cellcolor{red!30} 0.560 & \cellcolor{orange!30} 43min\\ 
			\bottomrule[1.2pt]
	\end{tabular}}
	\vspace{-0.2cm}
	\caption{Quantitative comparison results with different state-of-the-art (SOTA) methods on DTU without masks. \colorbox{red!30}{Best} and \colorbox{orange!30}{Second}-best are highlighted. In the last column, the optimization time for multi-view neural surface reconstruction is additionally compared. '-' stands for data not evaluated.}
	\vspace{-0.6cm}
	\label{table1}
\end{table*}

\subsection{Loss Function}
Similar to previous works, we use L1 loss to minimize the difference between the ground truth colors and the rendered colors:
\begin{equation}
	\label{eqn12}
	{{\mathcal{L}}_{rgb}}=\frac{1}{N}\sum\limits_{i=1}^{N}{|{{C}_{i}}-{{\widehat{C}}_{i}}|}
\end{equation}
where $N$ is the number of a batch rays. 

Eikonal loss is also added to regularize the SDF field.
\begin{equation}
	\label{eqn13}
	{{\mathcal{L}}_{eik}}=\frac{1}{MN}\sum\limits_{i,j=1}^{MN}{{{\left\| \nabla f({{p}_{i,j}})-1 \right\|}^{2}}}
\end{equation}
where $M$ is the number of sampled locations on each ray. 

Then the total loss is defined as:
\begin{equation}
	\label{eqn15}
	\mathcal{L}=({{\mathcal{L}}_{rgb}}+0.1{{\mathcal{L}}_{eik}})+({{\lambda }_{1}}{{\mathcal{L}}_{usdf}}+{{\lambda }_{2}}{{\mathcal{L}}_{bias}}+{{\lambda }_{3}}{{\mathcal{L}}_{pc}})
\end{equation}
The first two items are regular losses, and the last three items correspond to the three modules we proposed. In our experiments, we choose ${{\lambda }_{1}}=1$, ${{\lambda }_{2}}=1$, ${{\lambda }_{3}}=0.25$.
\section{Experiments}
\subsection{Experimental Setting}
\noindent\textbf{Datasets.} Following previous works, we use 15 scenes from the DTU dataset \cite{jensen2014large} for quantitative and qualitative evaluations, which include challenging cases, such as low-texture areas, thin structures, and specular reflection. Chamfer Distance (CD) is taken as the evaluation metric. Additionally, we also select several challenging scenes from the low-res BlendedMVS dataset \cite{yao2020blendedmvs} for qualitative comparisons.

~\\
\noindent\textbf{Network architecture.} Our network is built based on NeuS with light MLPs. The base SDF network is parameterized by a 4-layer MLP with 256 hidden units and the Bias network is parameterized by a similar 2-layer MLP. We also use a 2-layer MLP with 128 hidden units to construct the color network. To enhance learning efficiency, the popular InstantNGP \cite{muller2022instant} is utilized to encode the input position information of the three networks using the same parameters as MonoSDF \cite{yu2022monosdf}.

~\\
\noindent\textbf{Implementation details.} We use pre-trained TransMVSNet \cite{ding2022transmvsnet} on DTU to provide reliable depth map and point cloud for each view. To ensure fairness, we exclude the tested scenes from its original training set and retrain it on DTU. The time for depth prediction is about one second per image, thus the total preprocessing time is relatively short. The final point cloud contains about tens of millions of points. To demonstrate that our method does not rely on overly dense data, we downsample it to 2 million points by interval, i.e., sampling one point every $k$ points. This manner works well to balance efficiency while preserving the distribution characteristics of raw data. Furthermore, to facilitate the image projection in our Neural Projection module, we pick the point set generated by the current view to feed into the network at each iteration. In each batch, 512 rays and 1024 points are randomly sampled. We experiment on a single RTX 3090 GPU, optimizing for 30$k$ steps for around 43 minutes on each scene with a batch size of 1. Please refer to the supplementary material for more details.

\begin{figure}[t]
	\centering
	\includegraphics[width=\linewidth]{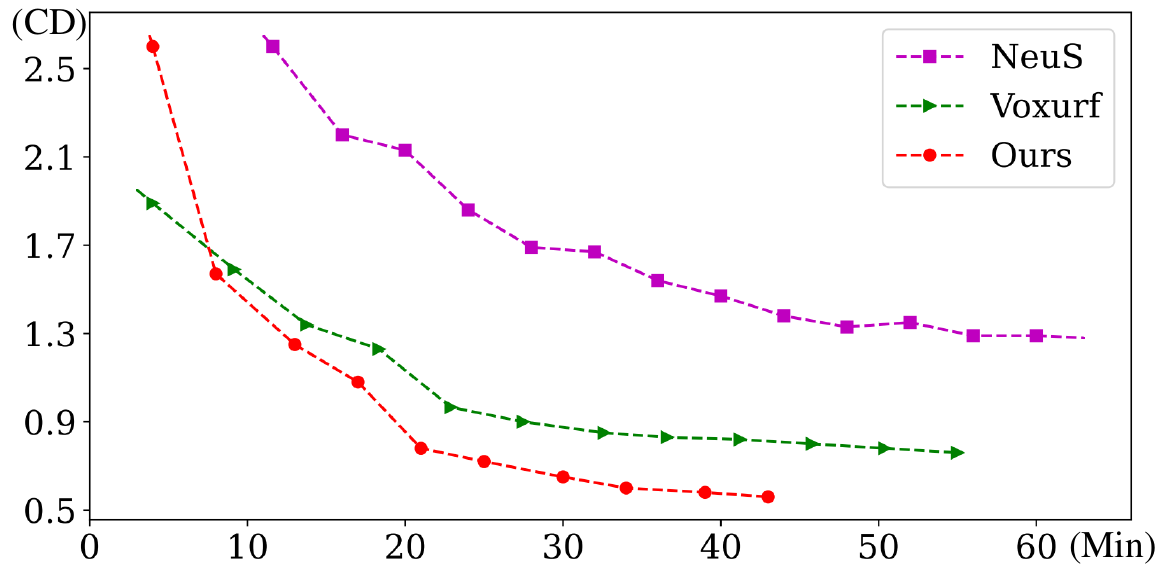}
	\vspace{-0.6cm}
	\caption{Convergence speed of different methods.}
	\vspace{-0.4cm}
	\label{fig7}
\end{figure}
\begin{figure}[t]
	\centering
	\includegraphics[width=\linewidth]{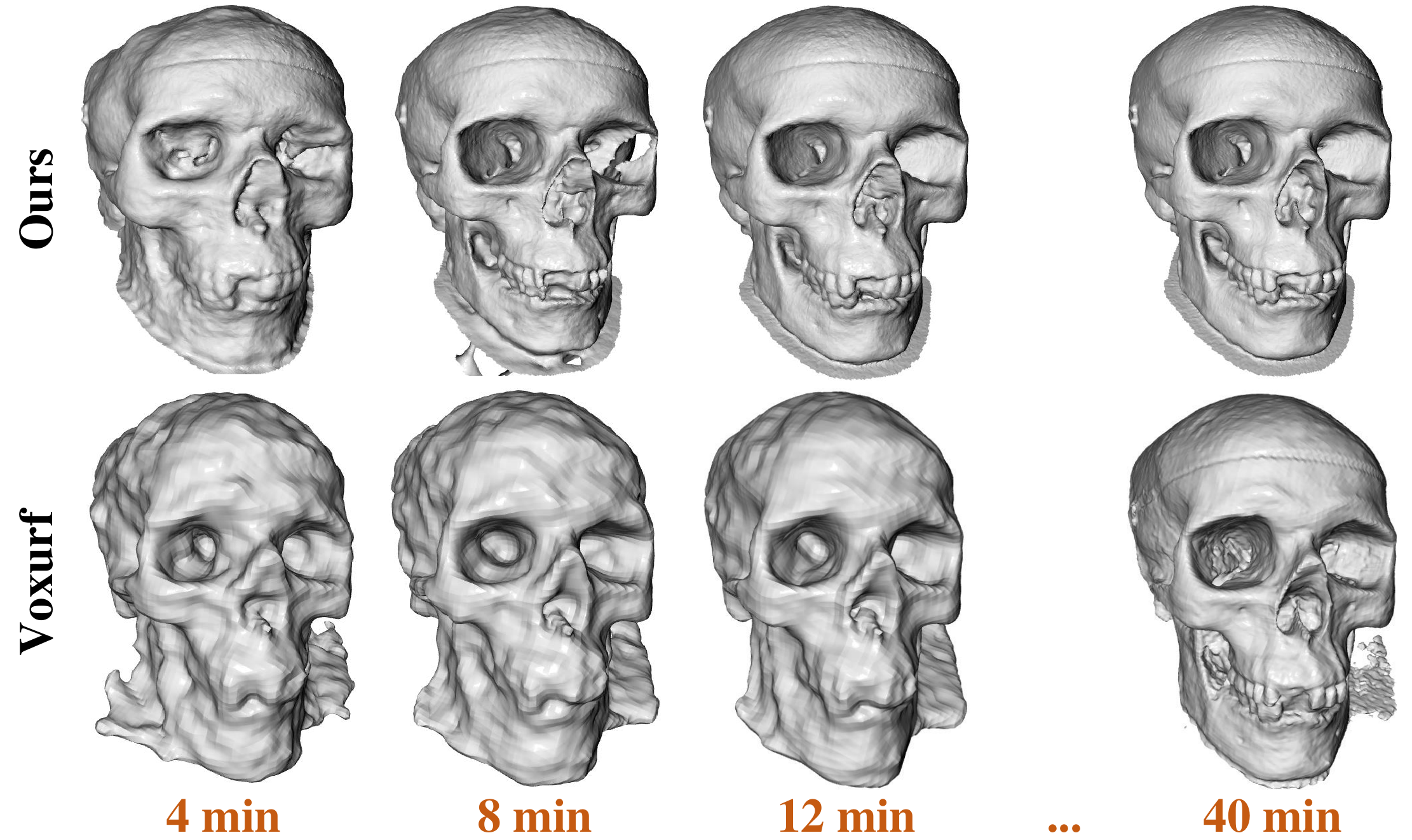}
	\vspace{-0.6cm}
	\caption{Reconstruction results at different optimization times.}
	\vspace{-0.4cm}
	\label{fig6}
\end{figure}

\noindent\textbf{Baselines.} We compare our method with \textbf{1)} sPSR \cite{kazhdan2013screened}, which is a classic geometric surface reconstruction method from point cloud. \textbf{2)} SIREN \cite{sitzmann2020implicit}, which is a popular learning-based implicit reconstruction method from point cloud. \textbf{3)} VolSDF \cite{yariv2021volume}, NeuS \cite{wang2021neus}, NeuralWarp \cite{darmon2022improving}, HF-NeuS \cite{wang2022hf}, DVGO \cite{sun2022direct}, Voxurf \cite{wu2022voxurf}, PET-NeuS \cite{wang2023pet} and NeuDA \cite{cai2023neuda}, which are neural surface reconstruction methods with volume rendering. \textbf{4)} MVSDF \cite{zhang2021learning}, Point-NeRF \cite{xu2022point}, MonoSDF \cite{yu2022monosdf}, RegSDF \cite{zhang2022critical} and TUVR \cite{zhang2023towards}, which use geometric priors to supervise neural surface reconstruction. In addition, COLMAP \cite{schonberger2016pixelwise} is also added.

\begin{table}[t]
	\centering
	\resizebox{\linewidth}{!}{
		\begin{tabular}{cccc}
			\toprule
			PET-NeuS \cite{wang2023pet} & HF-NeuS \cite{wang2022hf} & NeuralWarp \cite{darmon2022improving} & NeuS \cite{wang2021neus} \\
			\midrule
			30 GB & 16.21 GB & 9.28 GB & 6.84 GB \\
			\midrule
			Point-NeRF \cite{xu2022point} & MVSDF \cite{zhang2021learning} & Voxurf \cite{wu2022voxurf} & Ours \\
			\midrule
			23.30 GB & 18.85 GB & 23.55 GB & \textbf{4.88 GB} \\
			\bottomrule
	\end{tabular}}
	\vspace{-0.2cm}
	\caption{Comparison with different methods in terms of GPU memory usage.}
	\vspace{-0.4cm}
	\label{table2}
\end{table}

\begin{figure*}[t]
	\centering
	\includegraphics[width=\textwidth]{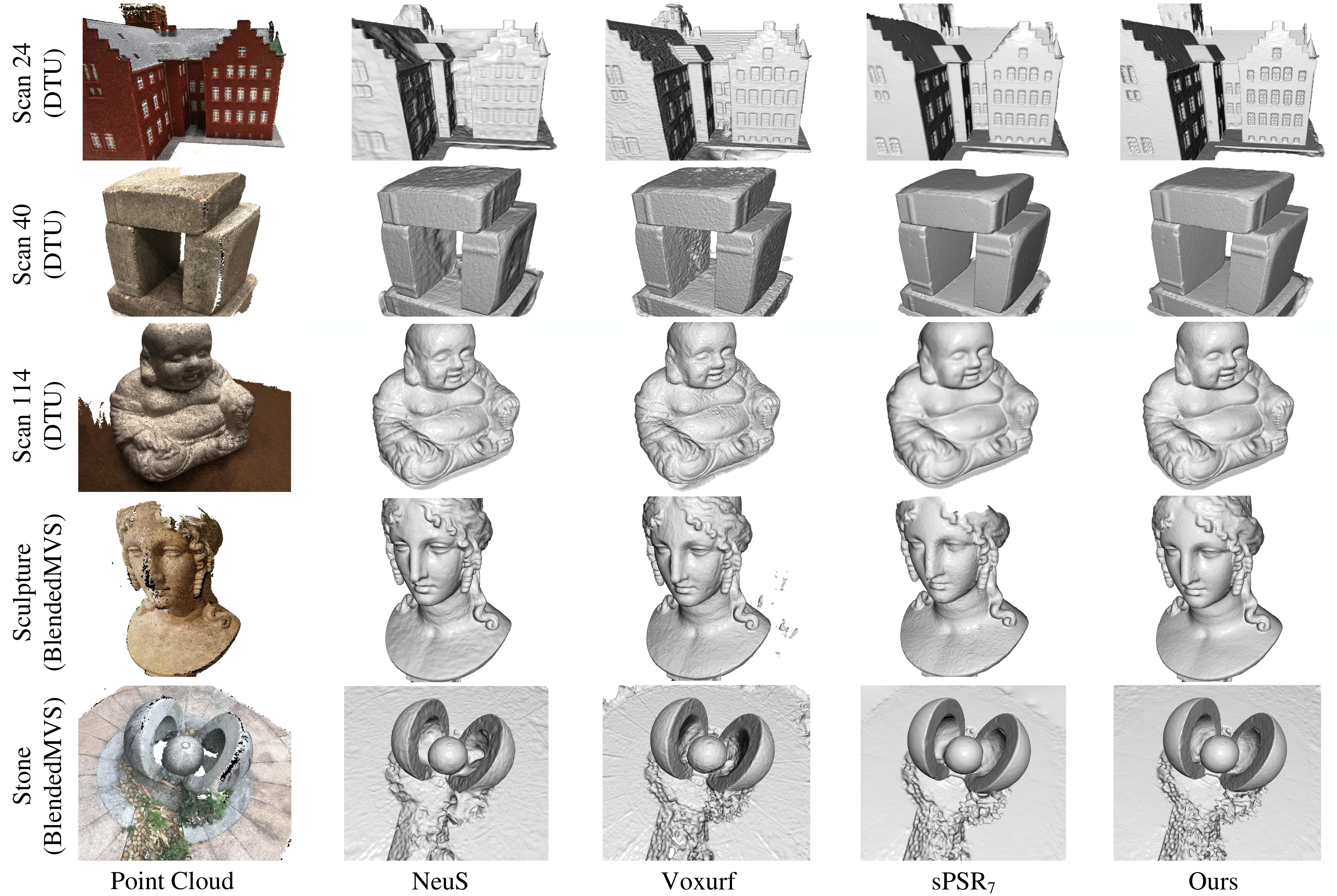}
	\vspace{-0.5cm}
	\caption{Qualitative comparison of our approach with methods from different categories on DTU and BlendedMVS.}
	\label{fig2}
	\vspace{-0.3cm}
\end{figure*}

\subsection{Comparisons}
As shown in Table \ref{table1}, we report the quantitative comparison results of our method with different methods on the DTU dataset without masks, including mesh quality and optimization time. Mesh quality evaluation results of all compared neural surface methods are derived from their original papers. Specially, the results of COLMAP are from the NeuDA paper \cite{cai2023neuda}, and the results of DVGO and Point-NeRF are from the Voxurf paper \cite{wu2022voxurf}. For sPSR, we report two results with trimming values set to 0 and 7. Although ${\text{sPSR}}_{7}$ scores higher than ${\text{sPSR}}_{0}$, the surface integrity is compromised. In addition, the mask cropping provided by NeuS is also applied to the results of sPSR and SIREN for fair comparisons. Importantly, both methods utilize undownsampled point clouds as input, making the comparison more challenging. In the comparison of accuracy with diverse baselines, our method shows significant advantages with a 33.3\% improvement over NeuS and a 12.5\% improvement over the SOTA method.

It can also be seen from Table \ref{table1} that the balance between accuracy and efficiency is still a challenge for current methods. DVGO \cite{sun2022direct} achieves a fast speed, but the reconstruction quality is not satisfactory. In contrast to other methods that require several hours for optimization, Voxurf and our method complete the task within one hour and reconstruct competitive results. Additionally, compared to Voxurf using multiple CUDA acceleration techniques for networks and sampling, our method achieves superior speed without relying on these, except for InstantNGP's CUDA implementation. The 11x speedup compared to NeuS and the 21.8\% speedup compared to Voxurf demonstrate the high efficiency of our method. As shown in Figure \ref{fig7} and \ref{fig6}, our method achieves fast convergence and a good reconstruction in a short time. As another advantage, it possesses a low GPU memory usage. Table \ref{table2} shows the comparison of our approach with different competitors in terms of resource usage. Even with lightweight networks, our approach still strikes a good balance between efficiency and accuracy, further demonstrating that the potential of point guidance is well exploited.

Figure \ref{fig2} presents a qualitative comparison of our method with other methods. Our approach combines the advantages brought by volume rendering and point modeling, resulting in improved reconstruction outcomes. Compared to other neural surface reconstruction methods, richer fine-grained details are characterized by our method, such as windows in Scan24 and cobblestones in Stone. Moreover, smoother surfaces are also reconstructed, like flat areas in Scan40 and curved ones in Scan114. Compared to sPSR, more complete surfaces and better noise robustness are presented by our method, such as the nose in Sculpture.

\begin{figure*}[t]
	\centering
	\includegraphics[width=\textwidth]{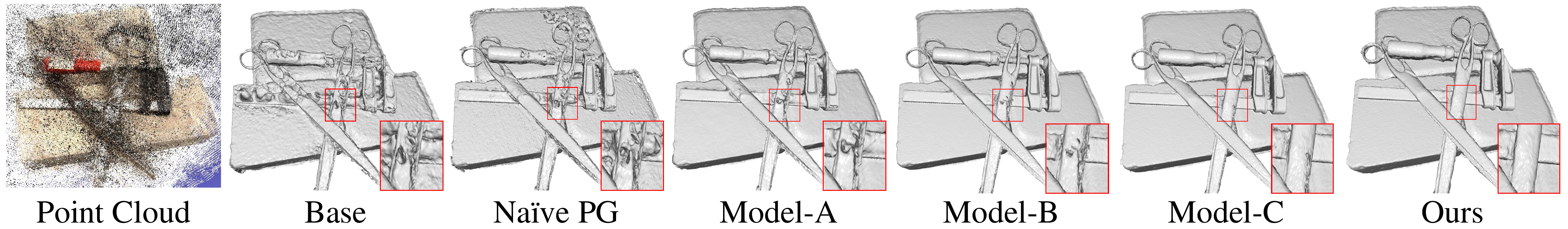}
	\vspace{-0.6cm}
	\caption{Visual results of different baselines when point clouds without geometric consistency filtering are taken as input.}
	\vspace{-0.5cm}
	\label{fig3}
\end{figure*}

\begin{table}[t]
	\centering
	\resizebox{\linewidth}{!}{
		\begin{tabular}{c|cccc|c}
			\toprule
			Mehtod & \; ${\mathcal{L}}_{sdf}$ \; & \; ${\mathcal{L}}_{pc}$ \enspace & \; ${\mathcal{L}}_{usdf}$ \; & \; ${\mathcal{L}}_{bias}$ \; & \; CD \; \\
			\midrule
			Base & & & & & 1.059 \\
			Naive PG & \checkmark & & & & 0.687 \\
			Model-A & \checkmark & \checkmark & & &0.649\\
			Model-B & & & \checkmark & &0.582\\
			Model-C & & \checkmark & \checkmark & &0.569\\
			Ours & & \checkmark & \checkmark & \checkmark & \textbf{0.560}\\
			\bottomrule
	\end{tabular}}
	\vspace{-0.1cm}
	\caption{Ablation study on DTU.}
	\vspace{-0.5cm}
	\label{table3}
\end{table}

\subsection{Analysis}
\noindent\textbf{Ablation study.} In order to evaluate the improvement brought by point guidance compared to the base framework and the effect of each module in our method, we conduct an ablation study on the DTU dataset, as presented in Table \ref{table3}. First, by removing the point modeling part of our pipeline, we only keep the volume rendering part as the Base model. Subsequently, we introduce the point cloud and a simple loss ${\mathcal{L}}_{sdf}$ as 'Naive PG', which is equivalent to ${\mathcal{L}}_{usdf}$ with ${{\widetilde{\sigma }}^{2}}\left( {{x}_{P}} \right)$ treated as a constant value in uncertainty estimation, to compare the effects brought by a naive point guidance. On this basis, we introduce 'Model-A', incorporating ${\mathcal{L}}_{pc}$ corresponding to the Neural Projection module, and 'Model-B', where ${\mathcal{L}}_{sdf}$ is replaced with ${\mathcal{L}}_{usdf}$ corresponding to uncertainty modeling. This allows us to evaluate the distinct effects of the two modules individually. The combination of the two modules is referred to as 'Model-C'. Finally, the Bias Network is added as our final model. It is evident that each module in our method contributes to higher-precision reconstruction, demonstrating the effectiveness of each module. More meaningfully, our point modeling is a universal module, not just for lightweight NeuS. The significant improvement over baseline not only illustrates the effectiveness of our point modeling module, but also illustrates the strong potential of point guidance.

\begin{table}[t]
	\centering
	\resizebox{\linewidth}{!}{
		\begin{tabular}{c|cccccc}
			\toprule
			Proportion & 0\% & 10\% & 15\%& 20\% & 25\% & 30\% \\
			\midrule
			CD & \textbf{0.560} & 0.565 & 0.571 & 0.576 & 0.588 & 0.602 \\
			\bottomrule
	\end{tabular}}
	\vspace{-0.2cm}
	\caption{Quantitative results of applying Gaussian noise to points of different proportions.}
	\label{table4}
	\vspace{-0.2cm}
\end{table}

\begin{table}[t]
	\centering
	\resizebox{\linewidth}{!}{
		\begin{tabular}{c|ccccc}
			\toprule
			Number & 5 Million & 2 Million & 1 Million & 0.5 Million & 0.1 Million \\
			\midrule
			CD & \textbf{0.560} & \textbf{0.560} & 0.565 & 0.572 & 0.575 \\
			\bottomrule
	\end{tabular}}
	\vspace{-0.2cm}
	\caption{Quantitative results with different number of points.}
	\label{table5}
	\vspace{-0.4cm}
\end{table}

\begin{figure}[t]
	\centering
	\includegraphics[width=\linewidth]{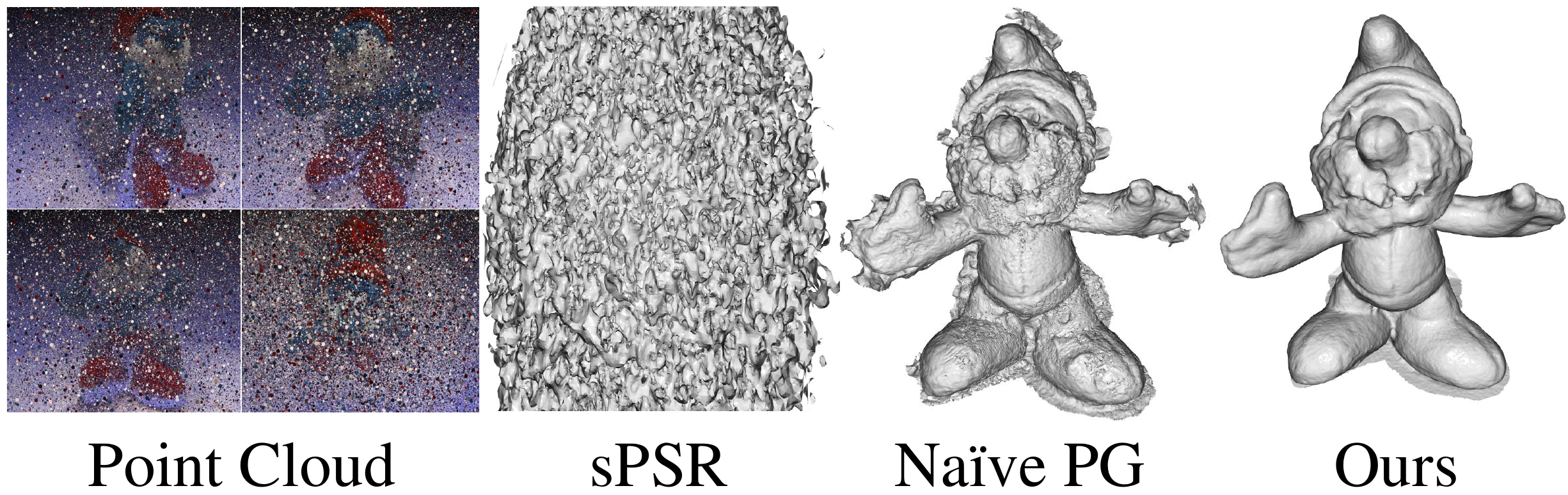}
	\vspace{-0.6cm}
	\caption{Reconstruction results with high-intensity artificial noise data. sPSR fails to work, while our method remains robust.}
	\vspace{-0.4cm}
	\label{fig4}
\end{figure}

~\\
\noindent\textbf{Noise resistance.} To test the robustness of our method to noisy point clouds, we perform experiments on two types of noisy data. The first type is a high-intensity noisy point cloud without any geometric consistency filtering during depth maps fusion, and the reconstruction results are presented in Figure \ref{fig3}. Additionally, the results of each baseline in Table \ref{table3} are also presented, further demonstrating the resistance of each module to high-intensity noisy data. To more clearly quantify noise intensity, we apply artificial noise to the point cloud as the second type of noisy data. We randomly select different percentages of points from the point cloud to add high-intensity Gaussian noise with a standard deviation of 50\%. As shown in Table \ref{table4}, as the noise intensity increases significantly, our CD value still maintains a small fluctuation, demonstrating strong robustness to noise. The visual results for high-intensity artificial noise are presented in Figure \ref{fig4}.

~\\
\noindent\textbf{Adaptation to density variations.} Adaptability to sparse data is also an important challenge for point-guided methods, reflecting the robustness to complex extreme cases. As shown in Table \ref{table5}, we downsample the raw points to different numbers to test the adaptability of our method for sparse data. It can be seen that our method remains a robust performance even as the point density significantly decreases. Satisfyingly, for sparse one hundred thousand points, we still maintain an excellent result.

~\\
\noindent\textbf{Limitations.} Reconstruction of non-visible regions remains a bottleneck in our approach. In these regions, point data usually cannot be reasonably generated, making it difficult to achieve effective surface guidance. This is also a common problem faced by current multi-view reconstruction. It would be a good direction to combine other generative models in future work. 

\section{Conclusion}
We propose PG-NeuS, a novel point-guided method for multi-view neural surface reconstruction, giving an effective solution to resist geometric ambiguity in the rendering and inherent noise in point cloud. Benefiting from effective point guidance, accurate and efficient reconstruction is achieved by our method, even using lightweight networks. Experiments demonstrate the superiority of our method in reconstruction accuracy, efficiency and robustness. 

{
    \small
    \bibliographystyle{ieeenat_fullname}
    \bibliography{main}
}
\end{document}